\documentclass[10pt,twocolumn,letterpaper]{article}

\usepackage{cvpr}
\usepackage{times}
\usepackage{epsfig}
\usepackage{graphicx}
\usepackage{amsmath}
\usepackage{amssymb}
\usepackage{amsfonts}
\usepackage{bbm}
\usepackage{xcolor}
\usepackage{subfig}
\usepackage{booktabs} % for professional tables
\usepackage{multirow}

\usepackage[pagebackref=true,breaklinks=true,letterpaper=true,colorlinks,bookmarks=false]{hyperref}

\cvprfinalcopy % 2 Uncomment this line for the final submission

\def\cvprPaperID{2} % *** Enter the CVPR Paper ID here

\ifcvprfinal\fi

\newcommand{\ZK}[1]{}

\begin{document}

%%%%%%%%% TITLE
\title{A probabilistic constrained clustering\\for transfer learning and image category discovery}

\author{Yen-Chang Hsu, Zhaoyang Lv\\
Georgia Institute of Technology, USA\\
%{\tt\small \{yenchang.hsu,zhaoyang.lv\}@gatech.edu}
\and
Joel Schlosser, Phillip Odom, Zsolt Kira\\
Georgia Tech Research Institute, USA\\
%{\tt\small {joel.schlosser,phillip.odom,zsolt.kira}@gatech.edu}
}

\maketitle
%\thispagestyle{empty}

%%%%%%%%% ABSTRACT
\begin{abstract}
Neural network-based clustering has recently gained popularity, and in particular a constrained clustering formulation has been proposed to perform transfer learning and image category discovery using deep learning. The core idea is to formulate a clustering objective with pairwise constraints that can be used to train a deep clustering network; therefore the cluster assignments and their underlying feature representations are jointly optimized end-to-end. In this work, we provide a novel clustering formulation to address scalability issues of previous work in terms of optimizing deeper networks and larger amounts of categories. The proposed objective directly minimizes the negative log-likelihood of cluster assignment with respect to the pairwise constraints, has no hyper-parameters, and demonstrates improved scalability and performance on both supervised learning and unsupervised transfer learning. 
\end{abstract}

%%%%%%%%% BODY TEXT
\section{Introduction}
Unsupervised clustering algorithms have found a number of applications ranging from data analysis, visualization, and importantly transfer learning and unsupervised image category discovery \cite{shu2018unseen,ovsep2017large,hsu2016deep}. Traditional methods that group data with a predefined metric, however, are very sensitive to the resulting choices or do not optimize the feature space jointly with clustering. If there is a complex nonlinear mapping between the input data and the desired semantic meaning of the clusters, the resulting clustering often cannot capture such meaning. In contrast, constrained clustering methods explicitly define the desired pairwise relationships as side information to the clustering algorithm. When paired with deep learning, the algorithm can learn the metric or feature representation to fit the desired relationships as well as optimize the assignments themselves.

Within this class of methods, the constraints often represent must-link/cannot-link or similar/dissimilar pairs (we use the latter in this paper). If a pair of data should be assigned to the same cluster, then it is a similar pair; otherwise it is dissimilar. The constraints could be regarded as an interface to inject prior knowledge for clustering. Such information can be obtained via human labeling (supervised or semi-supervised setting), spatial or temporal relationships e.g. in videos (unsupervised setting) \cite{Osep17arXiv}, or through a similarity metric learned on a different domain (transfer learning) \cite{Hsu18iclr}. Hence the constrained clustering formulation is a powerful concept that can bridge (semi-)supervised learning, unsupervised learning, and transfer learning.

A recent work \cite{Hsu18iclr} proposed a neural network-based constrained clustering objective in the form of a contrastive loss with KL-divergence. It demonstrated the transfer of learned pairwise similarity across tasks by clustering. Such a strategy can perform novel vision tasks, e.g. image category discovery with unlabeled examples of unseen categories. However, there are two critical challenges for this method: there is a degradation in the clustering performance when the chosen backbone network exceeds certain depth or when the number of clusters in the objective is larger (than a size of 10). These challenges prevent it from being applied to many real-world problems, which requires scalable solutions. Therefore, in this work we propose a new objective in the form of clustering likelihood to mitigate the mentioned issues.
\ZK{Might be hard to do space-wise but would be nice if intro included major contributions, or at least a sentence describing what we propose}
\section{The Constrained Clustering Likelihood}

This section considers the constrained clustering problem. Suppose $X=\{x_1,..,x_n\}$ denotes a dataset of $n$ instances, where each instance belongs to a set of unknown clusters $c\in\{1,2,..,k\}$. The set of constraints $S=S^+ \cup S^-$ contains two types of pairwise relationships, where $S^+$ is the set of tuples $(i,j)$ where $(x_i,x_j)$ is a similar pair, and $S^-$ contains the same for dissimilar pairs.

Similar to \cite{Hsu16iclrw,Hsu18iclr}, we define a neural network $f_\theta(x_i)=[p_{i,1},p_{i,2},..,p_{i,k}]^T$ that maps each sample $x_i$ to a categorical distribution, \ie $\sum_{c=1}^k{p_{i,c}}=1$, which is the probability of data $x_i$ belonging to cluster $c$. This is done by reinterpreting the softmax output of a neural network as outputting a probability distribution over cluster assignments, and a specialized loss function that utilizes the constraints is used to update $\theta$, the set of parameters in the neural network.

We introduce $y_i$ to represent $x_i$'s assigned cluster with probability $P(y_i=c|x_i)=p_{i,c}$. For a pair of data $(x_i,x_j)$, the probability of assigning both to the same cluster $c$ is:
\begin{equation*}
P(y_i=c,y_j=c|x_i,x_j)=P(y_i=c|x_i)P(y_j=c|x_j).
\end{equation*}
In the above formulation we impose an assumption that $y_i$ and $y_j$ are independent given their source data $x_i$ and $x_j$. For the simplicity of the representation, we omit the notation of conditioning since all of the following probabilities are conditioned on $X$.

Now we describe the probability of $(x_i,x_j)$ being a similar pair by marginalizing their joint distribution along the cluster index $c$:
\begin{equation} \label{eq:simi}
P(y_i=y_j) = \sum_{c=1}^k{P(y_i=c,y_j=c)} = f_\theta(x_i)^T f_{\theta}(x_j). \\
\end{equation}
Then the probability of $(x_i,x_j)$ being a dissimilar pair becomes straightforward:
\begin{equation} \label{eq:dissimi}
P(y_i \neq y_j) = 1-P(y_i=y_j).
\end{equation}
The constrained clustering likelihood $\mathcal{L}$ now can be defined as the product of all the probabilities of similar and dissimilar pairs:
\begin{equation} \label{eq:cclike}
\mathcal{L(\theta|S^+,S^-)} = \prod_{(i,j) \in S^+}{P(y_i=y_j)}\prod_{(i,j) \in S^-}{P(y_i \neq y_j)}.
\end{equation}
For finding $\theta$, we minimize the negative log-likelihood of equation \eqref{eq:cclike}, which is referred to as CCL and optimize this objective using stochastic gradient descent. For collecting the final cluster assignment, we only feed the data forward into $f_\theta$ again after the likelihood converged.

\section{Results}
We evaluate the proposed CCL and compare it with KL-divergence based constrained clustering loss (KCL) \cite{Hsu16iclrw,Hsu18iclr} in two settings using image datasets.

In the \emph{supervised} setting, we construct $S$ from ground-truth category labels. The motivation is to evaluate the upper-bound performance and explore scalability regarding the number of clusters and depth of the neural network, ablating the effect of noise. Tables \ref{deepnet} and \ref{supdatasets} demonstrate the superiority of CCL over KCL in terms of scalability. 

In the \emph{unsupervised transfer learning} (\ie image category discovery) setting, we follow the procedure for transferring across tasks \cite{Hsu18iclr}, which uses a learned similarity function as the constraint provider. The predicted constraints are noisy and therefore demonstrate the performance of our algorithm in a real-world setting. The scenario of unknown number of clusters is also evaluated by setting a large $k$ (\eg 100) in the network output. Table \ref{tab:corsstask_imagenet} shows the results of discovering 30 held-out categories in ImageNet \cite{deng2019imagenet}. In the case of unknown number of clusters, CCL (71.5\%) demonstrates a significant advantage over KCL (65.2\%). A similar trend also happens in the experiments of discovering characters in the Omniglot dataset (supplementary table \ref{tab:crosstask-omnilgot}) which includes comprehensive comparisons. Therefore we empirically conclude that CCL is a better clustering objective for these types of category discovery tasks.

\begin{table}
\centering
\caption{The accuracy on CIFAR10 with different loss functions and different neural network architectures. All training configurations, except the loss functions, are the same. The performance for Cross Entropy (CE) is the classification accuracy on test set and is regarded as the upper bound performance of supervised learning. For KCL and CCL, we give the clustering accuracy \cite{yang2010image} on the test set.}
\label{deepnet}
\begin{tabular}{@{}lc|ccc@{}}
\toprule
Networks           & CE & KCL     & CCL (ours)      \\ \midrule
LeNet              & 84.4 & 83.6  & \textbf{83.9}    \\ 
VGG8               & 89.4        & \textbf{89.6}  & 89.5    \\
VGG11              & 90.0        & 85.6  & \textbf{90.4}    \\
VGG16              & 91.1        & 21.9  & \textbf{91.3}    \\
PReActResNet-18    & 92.6        & 81.7  & \textbf{92.7}    \\
PReActResNet-101   & 93.0        & 21.9  & \textbf{93.3}    \\
\bottomrule
\end{tabular}
\end{table}

\begin{table}
\centering
\caption{The accuracy of supervised clustering with different number of clusters in the image datasets.}
\label{supdatasets}
\begin{tabular}{lccc|cc}
\toprule
Dataset  & \#class & Network & CE     & KCL   & CCL       \\ \midrule
MNIST    & 10      & LeNet & 99.4   & \textbf{99.5}  & 99.3     \\
CIFAR10  & 10      & VGG8 & 89.4   & \textbf{89.6}  & 89.5     \\
CIFAR100 & 100     & VGG8 & 64.1   & 44.3  & \textbf{64.0}     \\ 
\bottomrule
\end{tabular}
\end{table}

\begin{table}
\centering
\caption{Image category discovery on ImageNet. The values are the average of three random subsets in ImageNet\textsubscript{118}. Each subset has 30 classes. The "ACC" (clustering accuracy) has $K=30$ while the "ACC (100)" sets $K=100$. All methods use the features (outputs of average pooling) from Resnet-18 pre-trained with ImageNet\textsubscript{882} classification. NMI stands for normalized mutual information.}
\label{tab:corsstask_imagenet}
\begin{tabular}{lcccc}
\toprule
Method & ACC             & ACC(100)        & NMI            & NMI(100)       \\ \midrule
K-means & 71.9\%          & 34.5\%          & 0.713          & 0.671          \\
LSC    & 73.3\%          & 33.5\%          & 0.733          & 0.655          \\
LPNMF  & 43.0\%          & 21.8\%          & 0.526          & 0.500          \\ 
CCN-KCL    & 73.8\% & 65.2\% & 0.750 & 0.715 \\ \midrule
CCN-CCL    & \textbf{74.4\%} & \textbf{71.5\%} & \textbf{0.762} & \textbf{0.765} \\
\bottomrule
\end{tabular}
\end{table}

\subsubsection*{Acknowledgments}
This work was supported by the National Science Foundation and National Robotics Initiative (grant \# IIS-1426998).

\begin{table*}[]
\centering
\caption{Unsupervised cross-task transfer from $Omniglot_{bg}$ to $Omniglot_{eval}$ for discovering the characters in $Omniglot_{eval}$. The performance is averaged across 20 alphabets which have 20 to 47 letters. The ACC and NMI without brackets have the number of clusters equal to ground-truth.  The "(100)" means the algorithms use $k=100$, \ie one hundred outputs from the network. The characteristics of how each algorithm utilizes the pairwise constraints are marked in the "Constraints in" column, where metric stands for the metric learning of feature representation. }
\label{tab:crosstask-omnilgot}
\begin{tabular}{lcccccc}
\toprule
\multirow{2}{*}{Method} & \multicolumn{2}{c}{Constraints in} & \multirow{2}{*}{ACC} & \multirow{2}{*}{ACC (100)} & \multirow{2}{*}{NMI} & \multirow{2}{*}{NMI (100)} \\
                & Metric & Clustering &            &                              &                             &                              \\ \midrule
K-means \cite{macqueen1967some}        &     &      & 21.7\%    & 18.9\%  & 0.353   & 0.464  \\ 
LPNMF \cite{cai2009LPNMF}          &     &      & 22.2\%    & 16.3\%  & 0.372   & 0.498  \\
LSC \cite{chen2011LSC}            &     &      & 23.6\%    & 18.0\%  & 0.376   & 0.500  \\
ITML \cite{davis2007ITML}           & o   &      & 56.7\%    & 47.2\%  & 0.674   & 0.727  \\
SKMS \cite{anand2014SKMS}           & o   &      & -         & 45.5\%  & -       & 0.693  \\
SKKm \cite{anand2014SKMS}            & o   &      & 62.4\%    & 46.9\%  & 0.770   & 0.781  \\
SKLR \cite{amid2016SKLR}           & o   &      & 66.9\%    & 46.8\%  & 0.791   & 0.760  \\
CSP \cite{wang2014CSP}             &     & o    & 62.5\%    & 65.4\%  & 0.812   & 0.812  \\
MPCK-means \cite{bilenko2004MPCKMeans}     & o   & o    & 81.9\%    & 53.9\%  & 0.871   & 0.816  \\ 
CCN-KCL \cite{Hsu18iclr}            & o   & o    & 82.4\%    & 78.1\%  & 0.889   & 0.874  \\ \midrule
CCN-CCL (ours)     & o   & o    & \textbf{83.3\%}    & \textbf{80.2\%}  & \textbf{0.897}   & \textbf{0.893}  \\
\bottomrule                     
\end{tabular}
\end{table*}

{\small
\bibliographystyle{ieee}
\bibliography{egbib}
}

{\large\appendix{\textbf{Supplementary}}}

\section{Setup for supervised clustering}
\textbf{Network Architecture:} We use convolution neural networks with a varied number of layers: LeNet \cite{lecun1998gradient}; VGG \cite{simonyan2014VGG} and the PreActResNet \cite{he2016identity}. We also add a VGG8, which only has one convolution layer before each pooling layer, as the supplement between LeNet and VGG11. The number of output nodes in the last fully connected layer is the maximum number of clusters. We set it to the true number of categories for this section. Since the clustering objectives KCL and CCL both work on pairs of inputs, we insert a pairwise enumeration layer \cite{Hsu18iclr} between the network outputs and the loss function. Therefore the dense pairs in a mini-batch are all subject to the clustering loss.

\textbf{Training configurations:} All networks in this section are trained from scratch with randomly initialized weights. On the MNIST dataset, we use mini-batch size 100 with initial learning rate 0.1, which was dropped every 10 epochs by a factor of 0.1. We trained 30 epochs in total. On CIFAR10 we use the same setting except that the learning rate is dropped every 30 epochs and it is trained for 70 epochs in total. We use SGD for the optimization on MNIST and CIFAR10. For CIFAR100, the mini-batch size was 1000 and Adam was used as the optimizer with initial learning rate 0.001, which dropped every 70 epochs by a factor of 0.1. The training lasted 160 epochs.

\section{Setup for image category discovery}
We follow the unsupervised transfer learning procedures described in \cite{Hsu18iclr} to evaluate the clustering performance with noisy constraints. We use the original implementation and replace KCL by our CCL. We use the same training configurations, including the same similarity prediction function to make sure performance gain only by adopting CCL.  We follow \cite{Hsu18iclr} to use CCN for Constrained Clustering Network, and use CCN-KCL and CCN-CCL to differentiate the two approaches in the tables. Noted that the clustering accuracy is directly calculated on the target dataset instead of a holdout set since it is an unsupervised clustering setting.

\end{document}